\documentclass[letterpaper, 10pt]{IEEEtran}
\IEEEoverridecommandlockouts
\usepackage{graphicx}
\usepackage{amsmath, amssymb}
\usepackage{bbm}
\usepackage{cite}
\usepackage{mathtools}
\usepackage{algorithm}
\usepackage{algorithmic,booktabs}
\usepackage{caption}
\usepackage{subcaption}
\usepackage{bbold}
\usepackage{dsfont}
\usepackage{amsfonts}
\usepackage{tikz}
\usepackage{authblk}
\usetikzlibrary{arrows,automata,positioning}

\usepackage{xcolor}

\usepackage{tikz}
\usetikzlibrary{positioning, arrows.meta, shapes.geometric, fit, backgrounds, calc}
\usepackage{listings, url}
\usepackage{xcolor}
\lstdefinestyle{sigma}{
    basicstyle=\ttfamily\scriptsize,
    columns=fullflexible,
    keepspaces=true,
    showstringspaces=false,
    breaklines=true,
    frame=single,
    framerule=0.3pt,
    rulecolor=\color{black!35},
    backgroundcolor=\color{black!3},
    xleftmargin=3pt,
    xrightmargin=3pt,
    aboveskip=3pt,
    belowskip=3pt
}
\usepackage[T1]{fontenc}

\begin{document}
\title{Sigma-Hunter: A Domain-Specific Language Model for Threat Hunting and Detection Engineering}

\author{Kemal Davaslioglu}
\author{Sastry Kompella}
\affil{\normalsize Nexcepta, Gaithersburg, MD, USA 
\thanks{This work was supported in part by the U.S. Navy Small Business Innovation Research (SBIR) Program under Contract No. N6833525C0453. The authors gratefully acknowledge Dr. Walid Barnawi and Dr. Nathaniel Husted for their technical guidance, support, and valuable discussions throughout the project.}
} 

\maketitle
\thispagestyle{empty}
\begin{abstract}
Detection engineers must translate threat reports, forensic observations, and hunt hypotheses into precise, testable rules. General-purpose large language models (LLMs) can draft such rules, but often produce invalid YAML, incorrect log sources, unsupported fields, or overly broad detection logic. This paper presents \emph{Sigma-Hunter}, a domain-adapted LLM for analyst-assistive Sigma rule generation and threat hunting. We build an instruction-tuning dataset from 3,635 validated open-source Sigma rules, expanded into 7,663 question-answer and analyst-reasoning examples. Each source rule is assigned to a single train, validation, or test partition before this expansion, so no rule leaks across splits. We fine-tune a 7B Mistral model and a Phi-4 model with LoRA and score held-out rule generations on syntax, approximate field consistency, and a semantic judgment of detection logic, completeness, selectivity, and log-source alignment. Sigma-Hunter-Mistral scores 8.17 overall, against 7.88 for the strongest general-purpose baseline and 4.61 for untuned Mistral. Two findings stand out: domain adaptation enables a compact 7B model to perform competitively with larger general-purpose models on this structured task, and syntactic validity is a weak proxy for semantic rule quality, as several baselines emit well-formed YAML carrying weak detection logic. The adapted models run locally, which suits detection engineering in disconnected environments where analysts cannot reach hosted model services.
\end{abstract}


\section{Introduction}
Defensive cyber operations depend on converting threat information into detection logic that can be reviewed, tested, and deployed. Threat reports, endpoint alerts, malware analyses, and forensic observations may describe adversary behavior, but they rarely specify the log sources, event fields, filtering conditions, and false-positive controls required for an operational detection rule. This translation remains a time-consuming part of detection engineering, particularly in disconnected or air-gapped environments where analysts cannot rely on external model services.

Sigma provides a useful representation for this task. Its platform-agnostic YAML format expresses detection intent independently of a particular security information and event management (SIEM) backend, while retaining log-source requirements, detection conditions, severity, false-positive context, and optional MITRE ATT\&CK mappings~\cite{sigmahq_sigma,mitre_attack}. A valid Sigma rule can subsequently be translated or adapted for platforms such as Splunk, Elastic, or Microsoft Sentinel. However, portability does not guarantee quality: a rule may be syntactically valid yet operationally weak because it references unavailable fields, selects the wrong telemetry source, omits necessary filters, or encodes overly broad behavior.

Large language models (LLMs) offer a practical way to accelerate initial rule drafting, explanation, and refinement. Yet detection engineering is a high-precision structured-output task. General-purpose models frequently produce plausible rules that fail YAML validation, misuse Sigma fields, assume incorrect log sources, or require substantial analyst correction. We therefore investigate whether a compact model trained directly on validated Sigma artifacts can produce stronger draft rules than untuned general-purpose models while remaining suitable for local inference.

In this paper, we present a novel domain-adapted LLM called \emph{Sigma-Hunter} to assist analysts in Sigma rule generation and threat hunting. Sigma-Hunter is trained on question-answer and analyst-reasoning (QAR) examples derived programmatically from validated open-source Sigma rules. The key methodological choice is source-rule-level partitioning: each original Sigma rule is assigned to a single train, validation, or test partition before QAR expansion. This prevents different tasks derived from the same rule from appearing across training and evaluation sets.

This paper makes the following contributions:
\begin{enumerate}
\item We develop a reproducible pipeline that transforms validated Sigma rules into instruction examples for rule generation, explanation, ATT\&CK mapping, false-positive analysis, threat-hunting guidance, log-source reasoning, and refinement. The pipeline uses source-rule-level partitioning to keep every example derived from a rule within a single split.
\item We develop two domain-adapted models, Sigma-Hunter-Mistral (7B) and Sigma-Hunter-Phi4, using parameter-efficient fine-tuning and package them for local inference in disconnected environments.
\item We adapt multi-dimensional structured-output evaluation to Sigma generation, measuring syntax validity, approximate field consistency, detection logic, completeness, rule selectivity, and log-source alignment. Using this framework, the fine-tuned Mistral-7B model reaches 8.17 overall, compared with 7.88 for the strongest general-purpose baseline and well ahead of untuned Mistral (4.61), while demonstrating that syntax validity tracks semantic rule quality only loosely.
\end{enumerate}

The remainder of this paper is organized as follows. Section~\ref{section:background} reviews related work and the role of Sigma in detection engineering. Section~\ref{section:systemoverview} describes the Sigma-Hunter pipeline. Section~\ref{section:dataset_construction} presents dataset construction and rule-level partitioning, and Section~\ref{section:finetuning} details model fine-tuning and local deployment. Section~\ref{section:evaluation} reports the results. Section~\ref{section:conclusion} concludes the paper and outlines future work.

\section{Background and Motivation}\label{section:background}
\subsection{Related Work}

LLMs are being applied across cybersecurity, from threat-intelligence extraction to vulnerability analysis, offensive experimentation, and security-rule generation. Recent work shows that LLM agents can go beyond summarization to act: Fang \emph{et al.} autonomously exploited one-day vulnerabilities~\cite{fang2024llmagentsoneday}, Zhu \emph{et al.} extended this to teams of agents for zero-day scenarios~\cite{zhu2024teamsllmagentszeroday}, and Anthropic reported strong offensive and dual-use capabilities in a frontier model while arguing for matching defensive use~\cite{anthropic2026mythos}. These results concern offense and discovery rather than detection engineering, but they sharpen our motivation: as model-assisted offense improves, defenders need detection tooling that keeps pace and runs inside the enclaves where the relevant telemetry lives.

Other efforts focus on transforming cyber threat intelligence into structured artifacts. IntelEX uses LLMs for attack-level threat-intelligence extraction~\cite{intelex}, while LLMCloudHunter extracts detection rules from cloud-based CTI~\cite{llmcloudhunter}. More directly related, RulePilot proposes LLM-assisted security-rule generation~\cite{rulepilot}, while SigmaGen fine-tunes an LLM for ATT\&CK-aligned Sigma generation from threat intelligence~\cite{sigmagen}. CTI-REALM introduces a benchmark for detection-rule generation~\cite{ctirealm}, while GenTI applies related syntax-, semantic-, coverage-, and effectiveness-based evaluation to generated Snort, Suricata, and YARA rules~\cite{genti}. These works highlight the growing interest in LLM-assisted detection engineering and the need to evaluate schema adherence, rule correctness, field validity, and operational usefulness.

Sigma-Hunter extends this line of work through source-rule-level leakage control, matched tuned/untuned comparisons, quantitative Sigma-specific evaluation, and compact local deployment. Rather than proposing a new model architecture, it contributes a controlled domain-adaptation and evaluation pipeline for Sigma rule generation, with source-rule-level leakage prevention and multi-dimensional rule-quality assessment. Its focus on compact local deployment suits military cyber operations, where air-gapped environments and governance requirements limit the use of cloud-hosted LLM services.

\subsection{Detection Engineering and Sigma Rules}
Defensive cyber operations rely on timely and accurate detection logic to identify adversary activity across endpoint, network, identity, cloud, and application telemetry. In practice, the availability of threat intelligence does not automatically translate into deployable detections. A malware report, vulnerability disclosure, incident note, or adversary tradecraft description may identify relevant behaviors, but an analyst must still determine which telemetry sources are available, which event fields capture the behavior, what values or patterns should be matched, and how the rule should be scoped to avoid excessive false positives.

Sigma is a platform-agnostic format for expressing detection logic that can be translated to SIEM backends such as Splunk, Elastic, and Microsoft Sentinel. The \texttt{logsource} block specifies required telemetry, while \texttt{detection} defines selections, filters, and conditions. This intermediate representation is useful in military environments with heterogeneous and disconnected logging stacks.

Despite this portability, Sigma rule development remains a specialized task. A syntactically valid rule may still be operationally weak if it uses the wrong log source, invalid fields, noisy conditions, missing filters, or assumptions about telemetry that is unavailable in the target environment. Related detection formats, such as YARA \cite{yara} for file and malware pattern matching, follow a similar detection-as-code philosophy but target different artifacts and telemetry. In both cases, portability depends on the quality of the original rule, which motivates domain-specific assistance for generating higher-quality detection logic.

\subsection{Language Models for Detection Engineering}
LLMs have shown promise for cyber summarization, code generation, query drafting, incident triage, and threat-intelligence interpretation. In detection engineering, an LLM can potentially translate a natural-language behavior description into a draft rule, explain existing detection logic, identify relevant ATT\&CK techniques, suggest false-positive considerations, or propose hunting hypotheses. These capabilities are attractive because they can reduce the manual burden of converting threat information into deployable analytics.

General-purpose LLMs, though, are not built for detection-engineering constraints. Their outputs can look plausible while carrying invalid YAML, missing conditions, wrong log sources, hallucinated field names, or overly broad logic. Fine-tuning a compact model on validated detection artifacts narrows this gap: the model picks up recurring Sigma structures, log-source conventions, field names, condition patterns, ATT\&CK mapping conventions, and analyst-facing explanation styles.

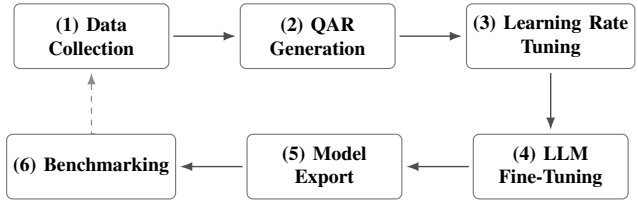
\begin{figure}[t!]
	\centering
	\resizebox{0.95\columnwidth}{!}{%
		\begin{tikzpicture}[
			>=Latex,
			font=\footnotesize,
			node distance=1.05cm and 1.25cm,
			stagebox/.style={
				draw=black!55,
				line width=0.55pt,
				rounded corners=3pt,
				minimum width=2.75cm,
				minimum height=1.15cm,
				align=center,
				fill=white,
				font=\bfseries
			},
			idx/.style={
				circle,
				draw=black!55,
				line width=0.5pt,
				fill=black!6,
				minimum size=0.52cm,
				inner sep=0pt,
				font=\scriptsize\bfseries
			},
			flow/.style={
				->,
				line width=0.7pt,
				draw=black!70,
				shorten >=2pt,
				shorten <=2pt
			},
			softflow/.style={
				->,
				dashed,
				line width=0.6pt,
				draw=black!45,
				shorten >=2pt,
				shorten <=2pt
			}
			]
			
			\node[stagebox] (s1) {(1) Data\\Collection};
			\node[stagebox, right=of s1] (s2) {(2) QAR\\Generation};
			\node[stagebox, right=of s2] (s3) {(3) Learning Rate\\Tuning};
			
			\node[stagebox, below=1.15cm of s3] (s4) {(4) LLM\\Fine-Tuning};
			\node[stagebox, left=of s4] (s5) {(5) Model\\Export};
			\node[stagebox, left=of s5] (s6) {(6) Benchmarking};
			
			\draw[flow] (s1) -- (s2);
			\draw[flow] (s2) -- (s3);
			\draw[flow] (s3.south) -- (s4.north);
			\draw[flow] (s4) -- (s5);
			\draw[flow] (s5) -- (s6);
			
			\draw[softflow]
			(s6.north) -| (s1.south);
			
		\end{tikzpicture}%
	}
	\caption{High-level end-to-end Sigma-Hunter pipeline that collects Sigma rules, generates QAR instruction examples, tunes the learning rate, fine-tunes the LLM, exports the model for local deployment, and benchmarks the resulting system.}
	\label{fig:pipeline}
\end{figure}

\section{Sigma-Hunter System Overview}\label{section:systemoverview}
Figure~\ref{fig:pipeline} shows the end-to-end Sigma-Hunter pipeline. The system converts open-source detection knowledge into a locally deployable domain-specific model through rule collection, parsing, validation, instruction-example generation, fine-tuning, export, and benchmarking.

Sigma-Hunter follows a staged pipeline for traceable deployment in disconnected cyber environments. The first stage collects Sigma rules from the SigmaHQ repository across emerging-threat, threat-hunting, operating-system, network, cloud, web, application, category, and identity directories. Each rule is parsed from YAML and stored with its raw content, source path, category, and metadata, preserving traceability from generated training examples back to the originating rule.

The second stage validates and cleans the collected rules. Rules are parsed as YAML and checked for required Sigma fields, including \texttt{title}, \texttt{logsource}, \texttt{detection}, and a detection \texttt{condition}. Additional checks are applied to optional fields such as \texttt{status}, \texttt{level}, \texttt{id}, \texttt{tags}, and \texttt{falsepositives}. Rules that are malformed, too short, too long, or contain problematic control characters are removed to reduce invalid syntax and corrupted formatting artifacts in training.

The third stage converts validated Sigma rules into QAR instruction examples for rule generation, rule explanation, ATT\&CK mapping, false-positive analysis, threat-hunting guidance, log-source reasoning, and rule refinement. For rule-generation examples, the input prompt includes a natural-language threat description and compact detection hints derived from the source rule's log-source and detection blocks. The target output is the original Sigma rule in valid YAML format. This design teaches the model to map analyst-facing descriptions and structured detection cues into complete Sigma rules.

The fourth stage performs supervised fine-tuning using parameter-efficient Low-Rank Adaptation (LoRA) \cite{hu2021lora}. The model is then exported to deployment-friendly formats, including GGUF/Ollama \cite{ollama}, for local inference, so that prompts, threat reports, logs, and detection artifacts stay within the operator-controlled environment.

Sigma-Hunter is therefore an analyst-in-the-loop capability, not an autonomous deployment system. In deployment, analyst-derived threat context and detection cues are provided to the model, which returns a candidate Sigma rule with a short explanation, likely log-source requirements, and false-positive considerations; and the analyst checks it against local telemetry and SIEM field mappings, runs it on historical data, tunes filters or allowlists, and only then promotes it through detection-as-code review. This preserves human accountability while shortening the path from a threat description to a reviewable detection artifact.

\section{Dataset Construction}\label{section:dataset_construction}
\subsection{Rule Collection and Validation}
We constructed the Sigma-Hunter dataset from open-source Sigma detection rules. In the collection stage, we cloned the SigmaHQ repository \cite{sigmahq_sigma} and extracted rules from directories covering emerging threats, threat hunting, operating-system telemetry, network telemetry, cloud services, web logs, applications, identity systems, and general rule categories. Table~\ref{tab:dataset_summary} summarizes the curated dataset. The rule collection process produced 3,635 source Sigma rules: 433 emerging-threat rules, 130 threat-hunting rules, and 3,072 generic detection rules. These source rules were expanded into 7,663 QAR instruction examples and partitioned, at the source-rule level, into 6,130 training, 766 validation, and 767 test examples. Our rule-generation evaluation uses the 188 rule-generation examples drawn from the 767-example test partition. Semantic judge metrics are computed over successfully judged outputs, with the number of valid judgments varying by evaluated model when a model failed to return a scorable rule or the judge response could not be parsed. The corpus is skewed toward generic detection rules, which make up 84.5\% of the source rules. We note that this may limit relative gains on emerging-threat and threat-hunting tasks.

\begin{table}[t!]
\centering
\caption{Sigma-Hunter dataset summary.}
\label{tab:dataset_summary}
\begin{tabular}{lr}
\toprule
\textbf{Item} & \textbf{Count} \\
\midrule
Source Sigma rules & 3,635 \\
Emerging-threat rules & 433 \\
Threat-hunting rules & 130 \\
Generic detection rules & 3,072 \\
Generated QAR examples & 7,663 \\ \midrule 
Training examples & 6,130 \\
Validation examples & 766 \\
Test examples & 767 \\
\bottomrule
\end{tabular}
\end{table}

Before generating instruction examples, candidate rules were cleaned and validated. Each rule was parsed as YAML and retained only if it contained a valid \texttt{title}, \texttt{logsource}, \texttt{detection}, and detection \texttt{condition}. Optional fields such as \texttt{status}, \texttt{level}, \texttt{id}, \texttt{tags}, and \texttt{falsepositives} were checked when present. Rules were removed if they failed YAML parsing, lacked required fields, contained invalid structural elements, exceeded length limits, or included problematic control characters. 

\subsection{QAR Instruction Example Generation}
Validated Sigma rules were transformed into QAR instruction examples using task-specific templates filled programmatically from source-rule fields. Table~\ref{tab:qar_tasks} summarizes the instruction-tuning task types and their corresponding target outputs. For the primary rule-generation task, the input prompt included: (1) the rule title and description as a natural-language threat behavior summary, (2) any available MITRE ATT\&CK tags, and (3) compact detection hints consisting of the condition expression, selection names, field names with modifiers such as \texttt{CommandLine|contains}, and representative field values. The prompt also included the intended log-source product, service, and category extracted from the rule's \texttt{logsource} block. The target response was the complete validated Sigma rule in YAML format. We therefore evaluate guided Sigma generation from partial analyst-supplied detection context, rather than unconstrained generation directly from threat reports.

Auxiliary templates were constructed for rule explanation, ATT\&CK mapping, false-positive analysis, and threat-hunting guidance. Rule-explanation prompts received the full YAML rule as input; ATT\&CK mapping prompts asked the model to identify techniques from rule tags and detection logic; false-positive analysis prompts asked for potential benign causes and mitigation filters; and threat-hunting prompts asked for a structured hunting hypothesis derived from the rule. All descriptions were derived from source-rule metadata rather than an external generative model, preserving traceability. Generated examples were validated for length, empty fields, and problematic characters to reduce malformed inputs and invalid YAML output patterns during training.

\begin{table}[t]
\centering
\caption{Instruction-tuning task types generated from Sigma rules.}
\label{tab:qar_tasks}
\begin{tabular}{p{0.34\linewidth}p{0.56\linewidth}}
\toprule
\textbf{Task Type} & \textbf{Target Output} \\
\midrule
Rule generation & Complete Sigma rule in valid YAML \\
Rule explanation & Purpose, log source, logic, severity, false-positive context \\
ATT\&CK mapping & Relevant tactics or techniques \\
False-positive analysis & Benign causes and analyst guidance \\
Threat-hunting guidance & Hunting hypothesis and follow-on steps \\
Log-source reasoning & Required telemetry source and fields \\
Rule refinement & Production-hardening guidance \\
\bottomrule
\end{tabular}
\end{table}

\subsection{Rule-Level Data Splitting}
A key concern is leakage between training and evaluation examples. Because multiple QAR examples can be derived from the same original Sigma rule, an example-level random split could place one task derived from a rule in training and another in testing, inflating performance. To prevent this, Sigma-Hunter uses source-rule-level splitting: original validated rules are assigned to train, validation, or test partitions before QAR expansion. As a result, all examples derived from a given source rule appear in only one split. The final split uses an 80/10/10 partitioning strategy at the source-rule level. Source-rule splitting prevents exact rule leakage, although closely related rule variants may remain across partitions; similarity-aware splitting is left for future work.

\section{Model Fine-Tuning and Deployment}\label{section:finetuning}
Sigma-Hunter was developed by fine-tuning compact instruction-tuned models for Sigma rule generation and detection-engineering assistance. Two variants were trained: Sigma-Hunter-Mistral and Sigma-Hunter-Phi4. These were evaluated against their corresponding untuned base models and additional general-purpose baselines. Training two variants tests whether domain-specific fine-tuning improves detection-engineering performance across model families rather than only for a single architecture.

Table~\ref{tab:finetune_config} summarizes the fine-tuning configuration. We used the \texttt{mistral-7b-instruct-v0.3-bnb-4bit} \cite{unsloth_mistral_7b_instruct_v03_bnb_4bit} model with approximately 7.29B total parameters. LoRA adapters were applied across all 32 transformer blocks, targeting the attention query, key, value, and output projections, as well as MLP feed-forward projections. This produced 41,943,040 trainable parameters, representing 0.575\% of the total model. The model and tokenizer used a vocabulary size of 32,768 tokens. Input examples were bounded by a maximum rule length of 4,000 characters, maximum description length of 500 characters, maximum response length of 3,500 characters, and tokenized sequence length of 2,048 tokens.

Training used supervised fine-tuning with LoRA \cite{hu2021lora} implemented through Unsloth and TRL. The model was loaded in 4-bit quantized form. Training used three epochs, per-device batch size of 4, gradient accumulation of 4, cosine learning-rate decay, warmup ratio of 0.03, 8-bit AdamW, and checkpoint selection by lowest validation loss. A learning-rate finder was used before full training to identify a stable operating region for each model-specific run. After training, LoRA adapters were saved, merged with the base model for inference, and exported as a GGUF/Ollama package for local deployment \cite{gguf}.

\begin{table}[t]
\centering
\caption{Fine-tuning configuration for Sigma-Hunter-Mistral.}
\label{tab:finetune_config}
\begin{tabular}{lp{0.50\linewidth}}
\toprule
\textbf{Parameter} & \textbf{Value} \\
\midrule
Base model & Mistral-7B-Instruct 4-bit \\
Training method & LoRA supervised fine-tuning \\
Trainable parameters & 41.94M (0.575\%) \\
Max sequence length & 2,048 tokens \\
Epochs / batch / grad. accum. & 3 / 4 / 4 \\
Optimizer / schedule & 8-bit AdamW / cosine \\
LoRA targets & Attention and MLP projections \\
Checkpoint selection & Lowest validation loss \\
Deployment export & GGUF/Ollama package \\
\bottomrule
\end{tabular}
\end{table}

\begin{figure}[t!]
	\centering
	\includegraphics[width=\columnwidth,trim={0 0 7in 0},clip]{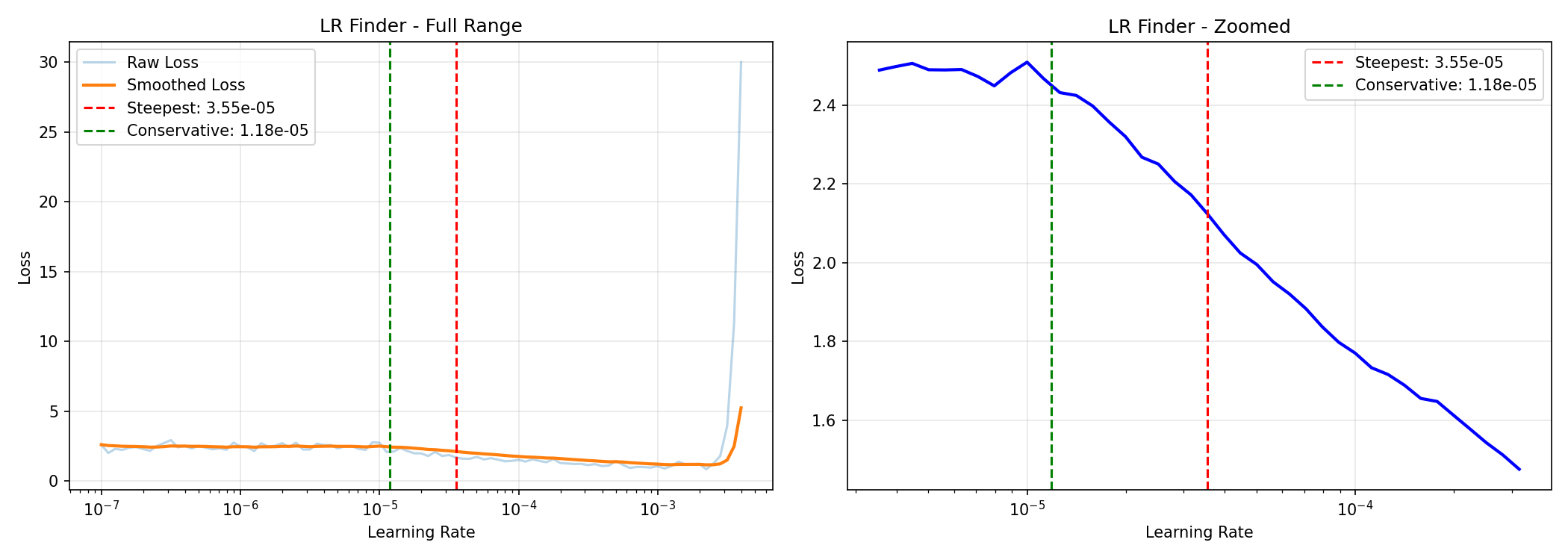}
	\caption{Different initial learning rates are evaluated for the Sigma-Hunter-Mistral model before the fine-tuning stage.}
	\label{fig:lr_finder}
\end{figure}

To select the learning rate for the Sigma-Hunter-Mistral run, we performed a learning-rate range test before full fine-tuning. As shown in Fig.~\ref{fig:lr_finder}, the loss remains relatively flat at very small learning rates and then begins to decrease more consistently as the learning rate increases. The steepest descent point occurs at approximately $3.55\times10^{-5}$, while a more conservative point is identified at approximately $1.18\times10^{-5}$. We used this range to guide learning-rate selection, favoring a stable value below the steepest-descent point to reduce the risk of divergence during LoRA fine-tuning \cite{hu2021lora}. The sharp loss increase at the upper end of the range indicates the onset of instability and was therefore avoided during full training.

\begin{table*}[t!]
	\centering
	\caption{Performance comparison across general-purpose and domain-specific LLMs for Sigma rule generation.} 
	\label{tab:model_results}
	\begin{tabular}{lccccccc}
		\toprule
		\textbf{Model} & \textbf{Syntax} & \textbf{Field} & \textbf{Detection} & \textbf{Completeness} & \textbf{Selectivity} & \textbf{Logsource Match} & \textbf{Overall} \\
		\midrule
		\texttt{qwen3:4b} & 0.02 & 0.03 & 3.29 & 2.54 & 3.30 & 7.02 & 3.83 \\
		\texttt{mistral:latest} & 3.84 & 1.73 & 3.56 & 3.92 & 4.17 & 8.36 & 4.61 \\
		\texttt{phi4:latest} & 7.89 & 6.07 & 6.20 & 5.81 & 6.28 & \textbf{9.98} & 6.82 \\
		\texttt{llama4:latest} & 9.09 & 6.54 & 6.98 & 6.36 & 6.96 & 9.93 & 7.36 \\
		\texttt{gpt-oss:20b} & 9.08 & \textbf{7.76} & 7.83 & 7.01 & 7.45 & 9.97 & 7.88 \\
		\texttt{gemma3:4b} & 7.64 & 1.07 & 3.83 & 4.48 & 4.69 & 9.94 & 5.28 \\
		\texttt{Sigma-Hunter-Mistral} & \textbf{9.29} & 7.73 & \textbf{8.04} & \textbf{7.59} & \textbf{7.95} & 9.95 & \textbf{8.17} \\
		\texttt{Sigma-Hunter-Phi4} & 8.17 & 7.01 & 8.01 & 7.42 & 7.90 & 9.49 & 8.03 \\
		\bottomrule
	\end{tabular}
\end{table*}

\section{Evaluations}\label{section:evaluation}
We report quantitative results only on the held-out rule-generation subset of the source-rule-level test partition. The remaining QAR task types contribute to training only when their source rules belong to the training partition; examples derived from validation or test rules are never used for training. Non-generation examples in the test partition are used only for qualitative analysis. The quantitative benchmark contains 188 held-out rule-generation examples. Each evaluated model received the same prompts. Generated outputs were post-processed to extract Sigma YAML by first searching for YAML code blocks and then for outputs beginning with common Sigma fields such as \texttt{title}, \texttt{logsource}, or \texttt{detection}. If no clear rule block was found, the raw response was retained for validation, allowing the benchmark to score both clean and imperfect outputs. Syntax and field scores are computed over generated outputs. Semantic scores are averaged over successfully judged outputs; Table~\ref{tab:model_results} reports the corresponding valid-judgment count.

The benchmark compared Sigma-Hunter-Mistral and Sigma-Hunter-Phi4 with several untuned general-purpose baselines: \texttt{qwen3:4b}, \texttt{mistral:latest}, \texttt{phi4:latest}, \texttt{llama4:latest}, \texttt{gpt-oss:20b}, and \texttt{gemma3:4b}. This comparison addresses two questions: whether domain-specific fine-tuning improves performance relative to a corresponding base model, and whether a compact domain-adapted model can perform competitively with larger general-purpose LLMs on a structured cyber task.

We evaluate generated Sigma rules using three metrics: syntax validation, approximate field consistency, and LLM-assisted semantic judging.

\emph{Syntax validation} measures if an output is a structurally valid Sigma rule candidate. Each rule is parsed as YAML and checked for five elements: a valid YAML dictionary, non-empty \texttt{title}, valid \texttt{logsource}, valid \texttt{detection}, and non-empty detection \texttt{condition}. Each element contributes two points to a 0--10 score. This metric identifies reviewable candidates, but does not assess semantic correctness.

\emph{Approximate field consistency} measures whether fields in the generated rule's \texttt{detection} block are appropriate for its declared log source. Extracted fields are compared with a curated dictionary for common telemetry categories, including Windows process creation and security, PowerShell, registry, file, network, Linux audit, macOS, cloud audit, proxy, and DNS logs. The resulting 0--10 score reflects the fraction of fields consistent with the declared log source. Because field availability varies across environments, this is an approximate signal rather than an authoritative SIEM schema validator.

\emph{LLM-assisted semantic judging} evaluates detection behavior beyond structure and field use. Given the task prompt, generated rule, and reference rule, a judge model scores detection logic, completeness, rule selectivity, log-source match, and overall quality on a 0--10 scale. Rule selectivity is a semantic judgment, not empirical alert selectivity measured on labeled telemetry. These dimensions assess behavioral correctness, coverage of reference indicators, detection scope, and telemetry alignment. Equivalent rules may receive high scores even when they do not textually match the reference.

We used \texttt{gpt-oss:20b} as a scalable semantic judge because it was the strongest locally deployable model available to us. As model-based judging is not independent expert validation, these scores should be interpreted as comparative semantic assessments; human detection-engineer validation remains future work.

Table~\ref{tab:model_results} summarizes performance on the rule-generation benchmark. \texttt{Sigma-Hunter-Mistral} achieved the highest overall score among the evaluated systems, increasing from 4.61 for untuned Mistral to 8.17 after Sigma-specific fine-tuning. Sigma-Hunter-Phi4 also improved over its untuned counterpart, increasing from 6.82 to 8.03. These matched-model gains are accompanied by improvements in detection logic, completeness, and selectivity, indicating that domain adaptation improves more than output formatting.

Syntax validity alone does not separate strong detection rules from merely well-formed outputs. Several models produced structurally valid rules and near-ceiling log-source match scores, yet differed materially in detection logic, completeness, selectivity, and field consistency. Sigma-Hunter-Mistral achieved the highest syntax, detection-logic, completeness, selectivity, and overall scores, while \texttt{gpt-oss:20b} scored slightly higher on approximate field consistency. Log-source match is best interpreted as a sanity check: it approaches the ceiling for most models and provides little separation among stronger systems. The very low \texttt{qwen3:4b} syntax and field scores reflect a different failure mode, as it rarely produced an extractable Sigma rule block.

\begin{table}[t]
	\centering
	\caption{Detection-logic excerpts from a held-out rule-generation example
		for local file reading using \texttt{curl.exe} with the \texttt{file:///}
		protocol handler.}
	\label{tab:worked_example}
	\footnotesize
	
	\textbf{Reference Rule}
	\vspace{2pt}
	\begin{lstlisting}[style=sigma]
		detection:
		selection_img:
		- Image|endswith: '\curl.exe'
		- OriginalFileName: 'curl.exe'
		selection_cli:
		CommandLine|contains: 'file:///'
		condition: all of selection_*
	\end{lstlisting}
	
	\vspace{3pt}
	\textbf{Sigma-Hunter-Mistral Output}
	\hfill \textbf{S: 10; F: 10; P: 10; O: 10}
	\vspace{2pt}
	\begin{lstlisting}[style=sigma]
		detection:
		selection_img:
		- Image|endswith: '\curl.exe'
		- OriginalFileName: 'curl.exe'
		selection_cli:
		CommandLine|contains: 'file:///'
		condition: all of selection_*
	\end{lstlisting}
	
	\vspace{3pt}
	\textbf{\texttt{gpt-oss:20b} Output}
	\hfill \textbf{S: 10; F: 10; P: 4; O: 5}
	\vspace{2pt}
	\begin{lstlisting}[style=sigma]
		detection:
		selection_cli:
		CommandLine|contains: 'file:///'
		condition: all of selection_*
	\end{lstlisting}
	
	\vspace{2pt}
	\raggedright\scriptsize
	\textit{Scores: S = syntax; F = field consistency; 
		P = rule selectivity; O = overall quality.}
\end{table}

Table~\ref{tab:worked_example} provides an illustrative held-out example of the gap between syntactic validity and semantic rule quality. In this case, \texttt{Sigma-Hunter-Mistral} reproduces the reference detection logic, including both the \texttt{curl.exe} executable constraint and the \texttt{file:///} command-line pattern, whereas \texttt{gpt-oss:20b} retains only the protocol-handler condition. Both \texttt{Sigma-Hunter-Mistral} and \texttt{gpt-oss:20b} produce syntactically valid rules with field-consistency scores of 10. However, \texttt{Sigma-Hunter-Mistral} preserves the reference rule's key constraints by requiring both \texttt{curl.exe} and the \texttt{file:///} command-line pattern. In contrast, \texttt{gpt-oss:20b} checks only for the protocol handler and can therefore match unrelated processes using the same URI scheme. This omission reduces its rule selectivity and overall scores to 4 and 5, respectively, while \texttt{Sigma-Hunter-Mistral} receives scores of 10 on both dimensions.

\section{Conclusion}\label{section:conclusion}
We presented a novel domain-adapted language model for analyst-assistive threat hunting and Sigma rule generation called Sigma-Hunter that transforms validated open-source detection rules into instruction examples, fine-tunes models for structured detection engineering, and packages the resulting capability for local inference. In our evaluations, Sigma-Hunter-Mistral achieved the highest mean overall score among the evaluated systems, while Sigma-Hunter-Phi4 also improved over its untuned baseline. The results show that domain adaptation improves reference-aligned semantic quality beyond syntactic correctness: rules that parse correctly and use plausible fields can still differ materially in detection logic, completeness, and rule selectivity. This distinction is important for defensive cyber operations in disconnected and air-gapped environments. Sigma-Hunter provides analysts with locally generated, reviewable detection drafts that preserve operational control over threat data and rule development. The intended workflow remains analyst-directed: generated rules are validated against local telemetry, tuned for false positives, and approved before deployment. Future work will incorporate human analyst validation of generated rules and extend Sigma-Hunter to backend-specific detection languages and operational schemas.

\bibliographystyle{ieeetr}
\bibliography{references}

\end{document}